\documentclass[conference]{IEEEtran}
\IEEEoverridecommandlockouts

\usepackage{cite}
\usepackage{amsmath,amssymb,amsfonts}
\usepackage{algorithmic}
\usepackage{graphicx}
\usepackage{textcomp}
\usepackage{xcolor}
\usepackage{url}
\usepackage{graphicx}
\usepackage{hyperref}
\usepackage{comment}

\usepackage[autostyle]{csquotes}
\def\BibTeX{{\rm B\kern-.05em{\sc i\kern-.025em b}\kern-.08em
    T\kern-.1667em\lower.7ex\hbox{E}\kern-.125emX}}
\begin{document}

\title{An LSTM-based Test Selection Method for Self-Driving Cars}

\author{\IEEEauthorblockN{Ali Güllü}
\IEEEauthorblockA{\textit{Institute of Computer Science} \\
\textit{University of Tartu}\\
Tartu, Estonia \\
ali.ihsan.gullu@ut.ee}
\and
\IEEEauthorblockN{Faiz Ali Shah}
\IEEEauthorblockA{\textit{Insitute of Computer Science} \\
\textit{University of Tartu}\\
Tartu, Estonia \\
faiz.ali.shah@ut.ee}
\and
\IEEEauthorblockN{Dietmer Pfahl}
\IEEEauthorblockA{\textit{Institute of Computer Science} \\
\textit{University of Tartu}\\
Tartu, Estonia \\
dietmar.pfahl@ut.ee}}

\maketitle

\begin{abstract}
Self-driving cars require extensive testing, which can be costly in terms of time. To optimize this process, simple and straightforward tests should be excluded, focusing on challenging tests instead. This study addresses the test selection problem for lane-keeping systems for self-driving cars. Road segment features, such as angles and lengths, were extracted and treated as sequences, enabling classification of the test cases as \enquote{safe} or \enquote{unsafe} using a long short-term memory (LSTM) model. The proposed model is compared against machine learning-based test selectors. Results demonstrated that the LSTM-based method outperformed machine learning-based methods in accuracy and precision metrics while exhibiting comparable performance in recall and F1 scores. This work introduces a novel deep learning-based approach to the road classification problem, providing an effective solution for self-driving car test selection using a simulation environment.

\end{abstract}


\section{Introduction}
Self-driving cars (SDCs), as key components of Cyber-Physical Systems, should be tested in simulation environments before being deployed in the real world in order to identify potential accidents, prevent financial losses, and not endanger human and animal lives \cite{b8}, \cite{b9}. Various simulation platforms, such as MATLAB/Simulink\footnote{\href{https://se.mathworks.com/solutions/automated-driving.html}{MATLAB/Simulink}}, BeamNG.tech\footnote{\href{https://beamng.tech/}{BeamNG.tech}}, Gazebo\footnote{\href{https://gazebosim.org/home}{Gazebo}}, CarSim\footnote{\href{https://www.appliedintuition.com/products/carsim}{CarSim}}, SUMO\footnote{\href{https://eclipse.dev/sumo/}{SUMO}}, PreScan\footnote{\href{https://plm.sw.siemens.com/en-US/simcenter/autonomous-vehicle-solutions/prescan/?srsltid=AfmBOoo7YraZL9AjT8SuWA6E-v6l7AIOJYPiUf7WeOYPd1098kPXBxeA}{PreScan}}, CARLA\footnote{\href{https://carla.org/}{CARLA}}, LGSVL\footnote{\href{https://hidetoshi-furukawa.github.io/post/lgsvl-simulator/}{LGSVL}}, and AirSim\footnote{\href{https://microsoft.github.io/AirSim/}{AirSim}}, have been developed for testing SDCs in the simulation world \cite{b10}, \cite{b12}.

Despite the benefits of simulation platforms, testing vehicles in these simulation environments is costly in terms of time. To minimize the testing costs, it would be helpful to identify and execute only those tests that are likely to cause failures and exclude overly simple tests from the test suite. To address this problem, we developed a test selection model for lane-keeping scenarios in SDCs. The proposed model, named \textbf{I}ntelligent \textbf{T}est \textbf{S}elector \textbf{for} \textbf{S}elf-\textbf{D}riving \textbf{C}ars (\textbf{ITS4SDC}), is based on a long short-term memory (LSTM) recurrent neural network \cite{b23} that focuses on identifying challenging driving scenarios to improve test efficiency by selecting failure-triggering roads as test cases. The developed model is also used in the \enquote{Self-Driving Car Testing Tool Competition}, organized as part of the 18th IEEE International Conference on Software Testing, Verification and Validation (ICST) 2025 \footnote{\href{https://conf.researchr.org/home/icst-2025}{ICST 2025}}.

To generate roads with diverse characteristics, the Frenetic road generation algorithm is used \cite{b1}. Frenetic is designed to create challenging roads that make it difficult for vehicles to stay within their lanes. It is one of the tools used in the SBST 2021 conference\footnote{\href{https://sbst21.github.io/}{SBST 2021}} to produce road types where vehicles most frequently went off-road. The Frenetic algorithm has also been utilized in other contexts, such as \cite{b4}, \cite{b5}, \cite{b18}.

To label generated roads as \enquote{PASS} (SAFE) or \enquote{FAIL} (UNSAFE), they are evaluated using the BeamNG.tech simulation platform, which offers realistic soft-body physics-based SDC testing. Its realistic physics, including friction forces and deformation during collisions, make it one of the most realistic simulation platforms for SDCs. 

A part of the labeled roads is then used as training data for models that aim at correctly classifying previously unseen roads as safe or unsafe. The rest of the labeled roads serve as the test set.

The evaluation results demonstrate that our LSTM-based model outperforms a set of state-of-the-art machine learning models in accuracy and precision while showing comparable performance in recall and F1 metrics.

The main contributions of this study are as follows:
\begin{itemize}
    \item Proposing a novel LSTM-based framework for efficient test selection for SDCs \cite{b24}.
    \item Providing a comprehensive comparison between LSTM and machine learning models for road classification.
    \item Preparation and public release of an open dataset for regression testing for SDCs \cite{b21}.
\end{itemize}

This paper is structured as follows. Section \ref{section_background} presents the background to the research presented in this paper. Section \ref{section_related_work} presents related work on testing SDCs for lane-keeping scenarios. Section \ref{section_experimental_design} covers the generation and labeling of test cases, as well as dataset preparation and model training. Section \ref{section_results} compares the proposed ITS4SDC model with SDC-Scissor, offering a detailed performance analysis. Section \ref{section_threats} summarizes the threats to validity. Section \ref{section_conclusion} highlights the advantages of the proposed model and outlines recommendations for future work.

\section{Background}\label{section_background}

\subsection{SDC Scenarios}
A scenario for SDCs represents a specific situation that the vehicle might encounter. An SDC scenario typically includes the vehicle itself, its environment, and surrounding factors such as roads, buildings, other vehicles or obstacles, and weather conditions \cite{b2}. These scenarios are generated in simulation environments to evaluate the behavior and performance of SDCs.

Among the fundamental scenarios for SDCs is lane-keeping, which has been extensively studied in previous studies \cite{b3}-\cite{b5}, \cite{b14}, \cite{b18}-\cite{b20}. This scenario focuses on one of the most basic safety functions of SDCs: preventing the vehicle from leaving the road while considering the geometry of the road. To ensure that the SDCs can adapt to various road geometries, it is essential to thoroughly test the lane-keeping system. For this purpose, the vehicle must be tested on roads with diverse characteristics, such as varying geometries, features, and lengths, and its behavior must be closely examined. However, it is impractical to test every possible road configuration due to the infinite number of potential features and geometries. Therefore, selecting or prioritizing challenging tests that are likely to cause the SDCs to fail or leave the road is crucial.

\subsection{SDC-Scissor Classification Approach}

Simulation tests are also time-consuming, and evaluating overly simple scenarios where the vehicle is almost guaranteed to succeed can be inefficient. To address the problem of selection \enquote{unsafe} test cases, the SDC-Scissor tool was proposed \cite{b4}. By analyzing road geometries and extracting relevant features, SDC-Scissor classifies roads as either \enquote{SAFE} or \enquote{UNSAFE}.

Another significant challenge in testing SDCs is generating roads that adequately test the limits of their lane-keeping systems. To address this, the SBST Tool Competition was organized, aiming to encourage the development of road generation tools designed to create challenging road scenarios for SDCs \cite{b17}.

\subsection{BeamNG.tech Simulation Environment}
The BeamNG.tech simulator was employed to test the lane-keeping system of SDCs. This simulator stands out among others due to its realistic physics engine, which closely models the unique dynamics of each vehicle, including skidding, traction, and braking behavior. These realistic physics provide a more accurate assessment of the performance of SDCs under diverse road conditions.

\subsection{Road Generation for SDC Testing}
\begin{figure}[htbp]
    \centering
    \includegraphics[width=0.45\textwidth]{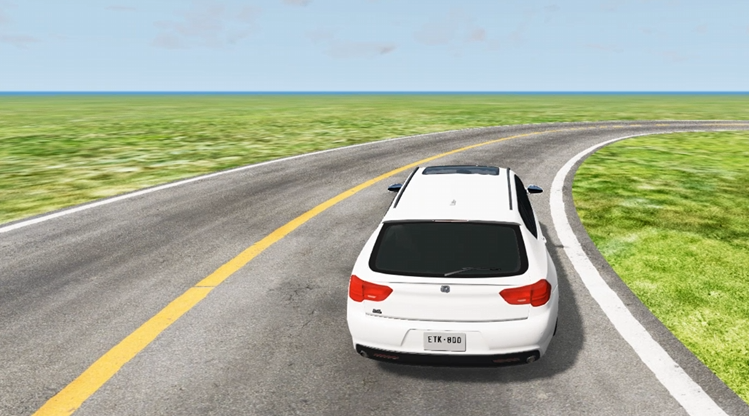}
    \caption{The vehicle stays in the lane (PASS).}
    \label{fig:vehicle_pass_ss}
\end{figure}

\begin{figure}[htbp]
    \centering
    \includegraphics[width=0.45\textwidth]{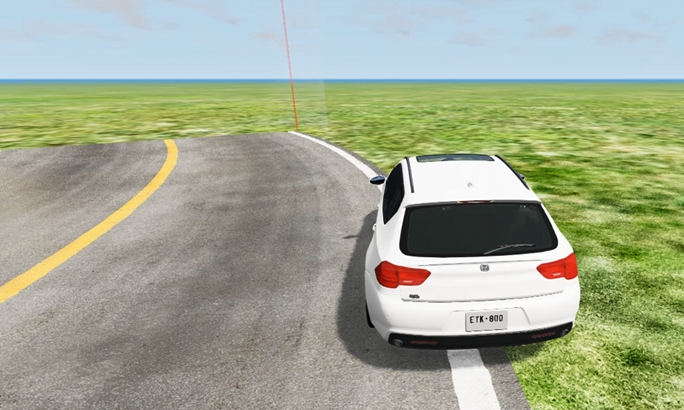}
    \caption{The vehicle goes off the lane while taking a turn (FAIL).}
    \label{fig:vehicle_fail_ss}
\end{figure}

The first step in testing the lane-keeping systems of SDCs is generating realistic roads. The (x, y) two-dimensional road coordinates are used to create road textures by extending the coordinates laterally on both sides to form a realistic asphalt surface in the simulation environment. Each test starts with a fresh simulation, and the simulation is terminated immediately if the vehicle either violates the defined out-of-bound (OOB) parameter (e.g., crossing the lane boundaries) or reaches the finish line. An example of a scenario where the vehicle successfully stays within the lane is illustrated in Figure \ref{fig:vehicle_pass_ss}. Conversely, Figure \ref{fig:vehicle_fail_ss} illustrates the scenario where the vehicle fails to navigate a curve and starts going off-road. The snapshot in Figure \ref{fig:vehicle_fail_ss} is captured just before the simulation is terminated.

\section{Related Work}\label{section_related_work}
Studies have been conducted on the problem of selecting safe and unsafe tests for SDCs. A review of these studies is provided in \cite{b2}, which compares several tools \cite{b13}-\cite{b18}. Among these tools, the SDC-Scissor tool specifically addresses the lane-keeping problem in SDCs \cite{b18}.

Birchler et al. (2023) conducted the most comprehensive study on lane-keeping systems for SDCs, focusing on road feature extraction and classification \cite{b5}. Road attributes (e.g., the number of left and right turns) and statistical metrics (e.g., segment standard deviation and median) were extracted using a segment-based strategy. These features were then classified using machine learning models such as Random Forest, Gradient Boosting, Support Vector Machine, Gaussian Naïve Bayes, Logistic Regression, and Decision Tree.

However, their approach relies on the assumption that statistical or behavioral differences exist between \enquote{FAIL} and \enquote{PASS} roads, which machine learning algorithms can detect by identifying distinct patterns.

However, this research considers the road as a sequence and leverages LSTM networks to address its sequential nature for classifying \enquote{SAFE} and \enquote{UNSAFE} test cases.

\section{Development of the ITS4SDC Model}\label{section_experimental_design}

This section describes the simulation environment, the autonomous driving of the vehicle in the simulation environment, and the behaviors of the SDCs. Additionally, it outlines the model’s step-by-step development processes. The feature extraction process from road coordinates will be addressed, followed by the training and evaluation of the model.

\subsection{Configuration of the Simulation Environment}

BeamNG.tech is used as a simulation environment. The two-dimensional map is used for the vehicle, and the vehicle does not perform any ascending or descending movements along the z-axis. The vehicle selected by the simulation is the ETK-800 model, which was also chosen by previous SDC test selection studies \cite{b4}-\cite{b7}. The BeamNG.tech simulation environment includes a variety of vehicles \cite{b22}. However, each vehicle's road grip varies due to differences in components such as the suspension system, tire characteristics, weight distribution and center of gravity, differential type, Traction Control System (TCS), Anti-lock Braking System (ABS), and Electronic Stability Control (ESC) units. For example, a turn that one vehicle can take at 50 km/h may not be possible for another vehicle at the same speed.

The ETK-800 model vehicle has been selected for conducting tests in the BeamNG.tech simulator. The same vehicle was also chosen by previous SDC test selection studies \cite{b4}, \cite{b7}. In BeamNG.tech, vehicles can be controlled either manually or autonomously. For autonomous control, BeamNG.AI, developed by BeamNG.tech, or a custom-developed driver can be utilized. The BeamNG.AI is capable of performing various tasks such as moving on random roads, pursuing a target vehicle, avoiding a target vehicle, maintaining a safe following distance with a target vehicle, and tracking a predefined path. For this study, the configuration is set to make the vehicle follow a given path. the AI driver drives the vehicle based on specific driving parameters, including maximum speed (MaxSpeed), risk factor (RF), and out-of-bound tolerance (OOB). 

The SDCs drive at the MaxSpeed if there are no obstacles and the road is safe and straight. The RF defines the aggressive level of the SDCs. If RF is high (e.g., 2.0), the vehicle maintains a high speed and takes significant risks, often causing it to lose stability on sharp curves. When the RF is moderate (e.g., 1.5), the vehicle tends to approach the edges of the road unless the curve is too sharp, in which case it slows down. A low-risk factor (e.g., 1.0) minimizes risk, allowing the vehicle to maintain maximum speed only on completely straight roads. Even on slightly curved roads, the vehicle slows down significantly before entering the curve. OOB tolerance defines the permissible extent (as a percentage) of the vehicle's deviation from the road or lane markings before marking the test as a failure. If the simulation is required to stop and the test to be labeled as \enquote{FAIL} when 50\% of the vehicle goes off the road, the OOB value is set to 0.5. The maximum speed is set to 120 km/h, the risk factor to 1.5, and the OOB tolerance to 0.5, following the parameters used in the ICST 2025 SDC Testing Tool Competition Track. The BeamNG simulator processes the test cases (roads) using the Test-Execution framework provided by SDC-Scissor \cite{b6}.

\subsection{Dataset Preparation}\label{subsection_dataset_preparation}

To test SDCs in a simulation environment, the Frenetic tool, which uses a genetic algorithm, is utilized to generate roads with varying characteristics such as the number of curves, turns, and total distance \cite{b1}. However, not all roads generated by the Frenetic algorithm are valid. A valid road cannot contain self-intersections or partially overlapping paths \cite{b3}. During preprocessing, such invalid roads were removed from the Frenetic-generated dataset, leaving only valid roads to be used as test cases.

In some test cases, when the target coordinate is positioned very close to the starting point, the AI Driver occasionally attempts to reach the target point by reversing or making an abrupt maneuver at the beginning of the test. While these cases were labeled as \enquote{PASS} by the simulator, they were excluded from the test suite as the vehicle did not follow the intended path. Ultimately, a dataset of 10,000 test cases, each labeled as \enquote{PASS} or \enquote{FAIL} based on the road-following outcomes, is prepared and referred to as \enquote{Dataset-1}. In Dataset-1, the vehicle failed to follow the road in 3,853 test cases, either leaving the road and entering an off-road area or crossing into another lane during a turn. The vehicle successfully completed the road in 6,147 test cases.

\subsection{Building the ITS4SDC model Using LSTM}

\subsubsection{Identifying road features} An initial analysis of Dataset-1 was conducted to examine the characteristics and statistical differences between roads labeled as \enquote{FAIL} and \enquote{PASS}, as in \cite{b4}. These characteristics and analyses focused on segment-based road examination, where each segment is defined as the line connecting two consecutive road coordinates. Features such as the total number of left/right turns, average turning angles, and turning radius were considered.
However, the segment-based analysis revealed no significant feature distinctions between \enquote{FAIL} and \enquote{PASS} test cases even though the dataset included 3,853 \enquote{FAIL} tests. The inability to differentiate directly based on road characteristics is affected by the risk factor considered by the AI Driver of the vehicle. The following examples illustrate the challenges of directly separating \enquote{FAIL} and \enquote{PASS} cases based on road features:

\begin{itemize}
\item  On roads with very sharp curves, the significant angle changes and increased number of turns often can be considered as a challenging path. However, if the curve is located at the beginning of the road, the vehicle can navigate it before gaining high speed and successfully completing the route.
\item In the case of two consecutive curves, the vehicle successfully passes the first non-sharp curve, loses some speed, and safely enters and exits the second curve. However, the presence of multiple curves causes significant fluctuations in the angle statistics of the mentioned road.
\item On a road with a low turning angle, if the vehicle starts on a straight path, it fails to navigate the curve and leaves the road due to gained speed during the straight part of the road.
\item A vehicle traveling on a straight road successfully navigates a curve located very close to the target coordinate. However, the reason for slowing down is not the curve itself, but the proximity of the target coordinate immediately after the curve, as the AI Driver considers the entire road.
\end{itemize}

These observations indicate that the road characteristics alone do not determine whether the vehicle will go off the road; the driver's behavior is equally important. Initial attempts were made to classify roads using machine learning models based on their features, as in \cite{b4},\cite{b5},\cite{b18}. However, the distribution of road features showed no significant distinction across Dataset-1, and the apparent separation observed with a small dataset diminished as the dataset grew to 10,000 test cases.

In segment-based road analysis, it is essential to consider road features as a sequence. Two primary features of the road have been examined:
\begin{itemize}
    \item Segment Angle: The angular displacement formed between a segment defined by two consecutive road positions and an adjacent segment.
    \item Segment Length: The length of a segment is defined by an Euclidean distance between two consecutive road positions.
\end{itemize}
The segment angle provides information about curves on the road. Sudden changes in segment angles represent sharp turns. However, the angle alone is insufficient due to variations in the distances between consecutive road positions, which depend on the road generation or interpolation algorithms. Since the duration of a vehicle’s turn is determined by the length of the corresponding segment, segment length has been considered as the second feature.

On the other hand, considering only segment length as a sequence lacks information about curves and, therefore, fails to provide meaningful information about the safety of the road.

\begin{figure}[htbp]
    \centering
    \includegraphics[width=0.45\textwidth]{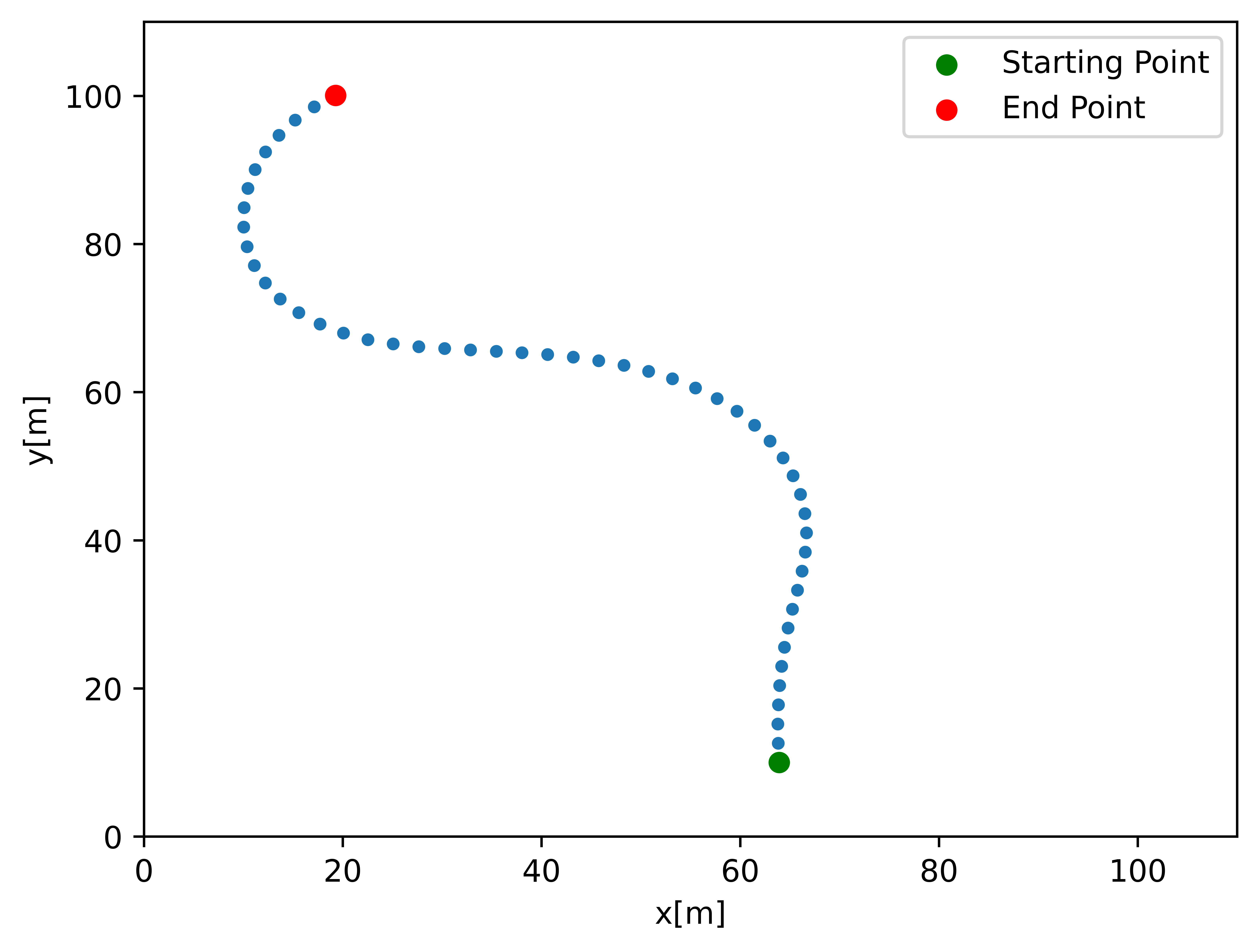}
    \caption{The road coordinates of a test case.}
    \label{fig:example_road_points}
\end{figure}

\begin{figure}[htbp]
    \centering
    \includegraphics[width=0.45\textwidth]{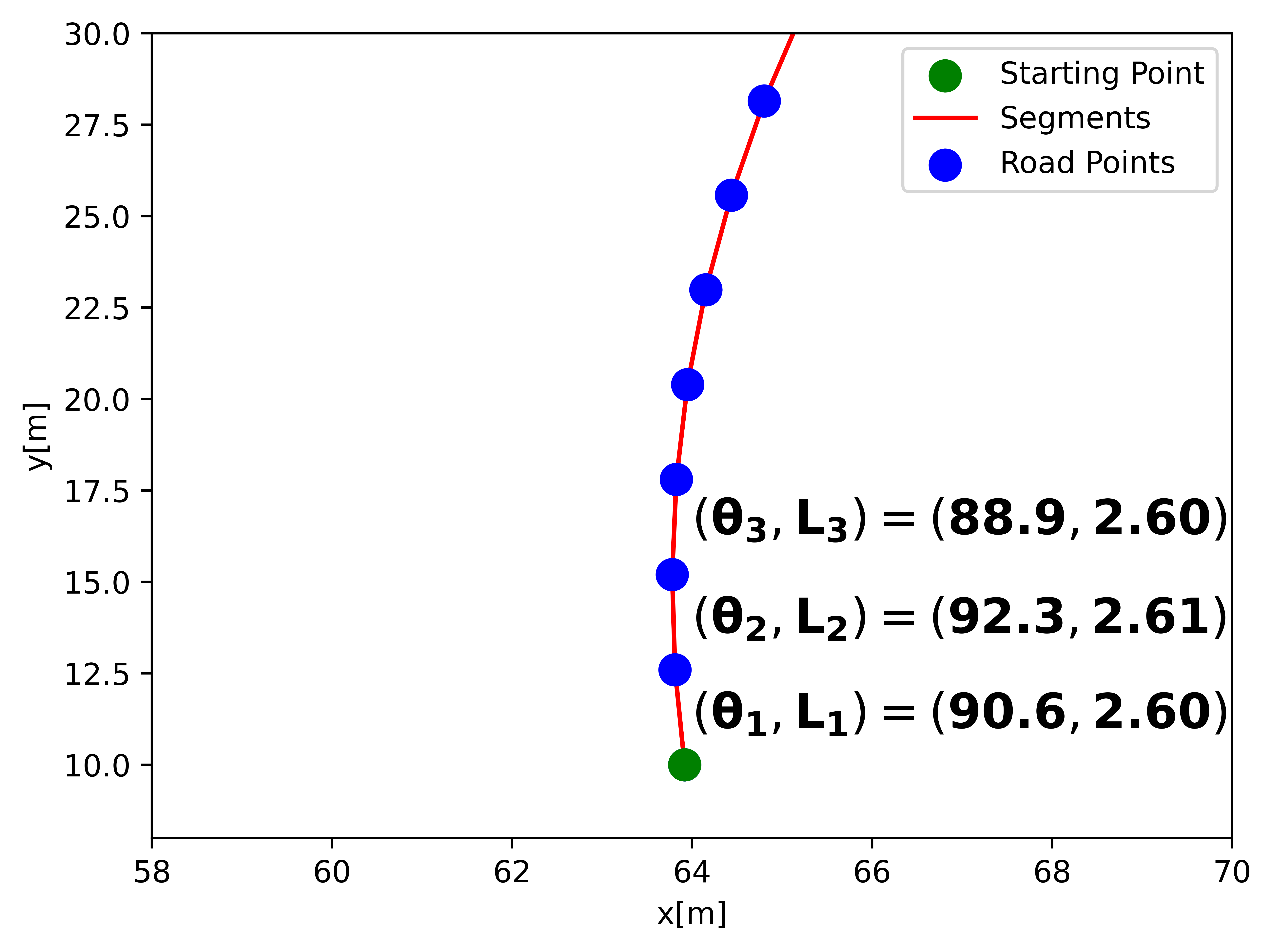}
    \caption{Angle and length pairs corresponding to the road coordinates}
    \label{fig:raw_features_of_the_road}
\end{figure}

\begin{figure}[htbp]
    \centering
    \includegraphics[width=0.45\textwidth]{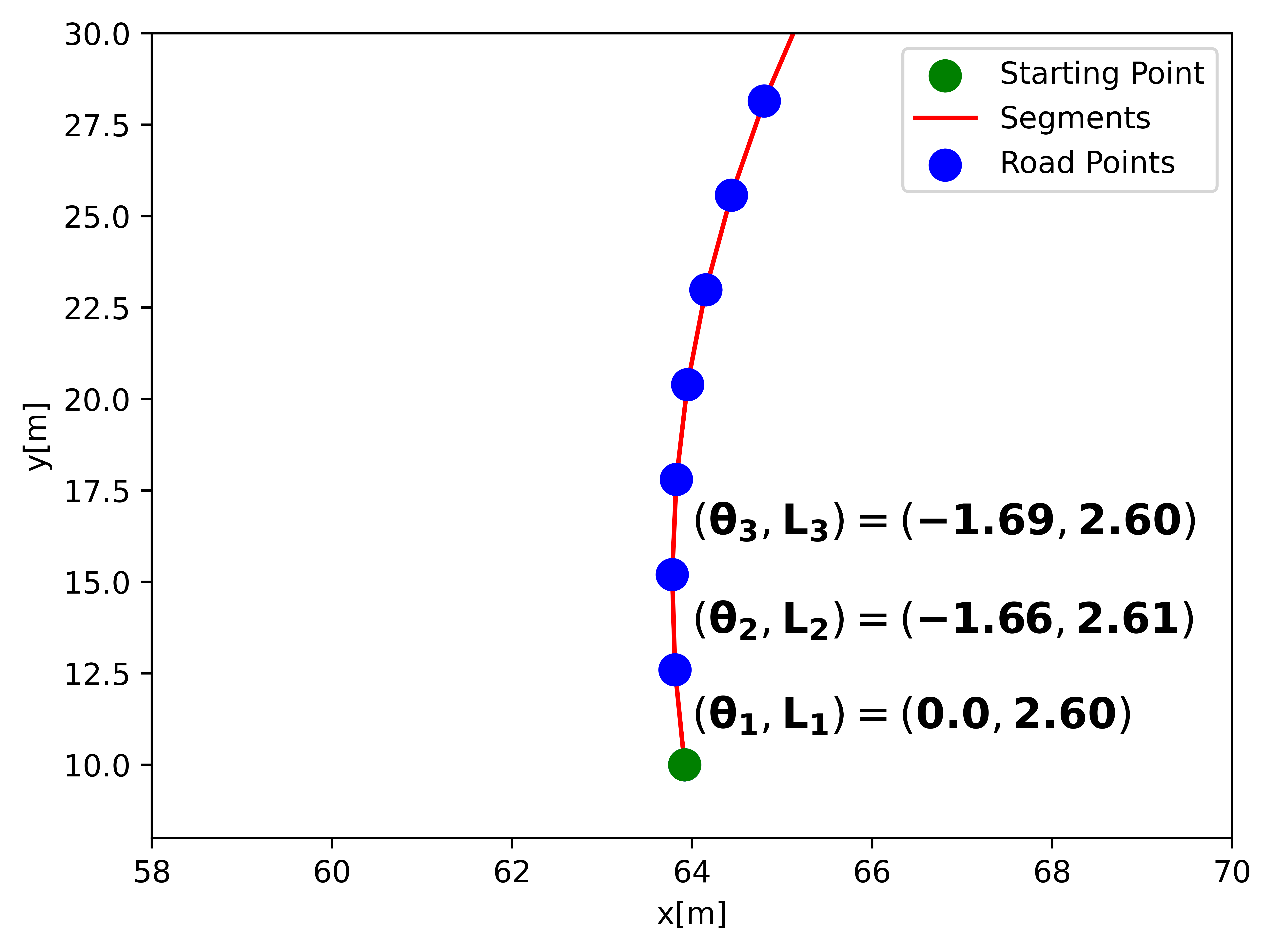}
    \caption{Calculated angle information relative to previous reference segments.}
    \label{fig:adjusted_angles_of_the_road}
\end{figure}

The vehicle's ability to stay on the road depends not only on the angle between two consecutive segments but also on the sequential behavior across multiple segments. This makes the problem a sequence classification task, as the sequence characteristics are influenced by both the $N$-th segment of the road and its neighboring segments, including $N+r$ elements \[r = (..., -2, -1, 0, 1, 2, 3, ...)\] where $r$ denotes the relative position of neighboring segments (e.g., past, present, or future). To address this, LSTM models, which are well-suited for sequence classification tasks, were employed \cite{b11}. The classification labels were binary: 0 for \enquote{FAIL} and 1 for \enquote{PASS}.

Initially, for a $(2, N)$ road coordinate array, such as $\{(x_1, y_1), (x_2, y_2), \ldots, (x_{N}, y_{N})\}$, the feature extraction is derived \((\textit{segment}_{\text{angle}}, \textit{segment}_{\text{length}})\). The road data of a test case is presented in Figure \ref{fig:example_road_points}. Starting from the initial point, $N$ points constitute the road coordinates. As a result, a feature vector of size $(2, N-1)$ is obtained. This feature vector, which captures sequential road characteristics such as changes in angle and segment length, serves as the input vector for the LSTM layer.

Angles are calculated relative to the two-dimensional Cartesian coordinate system. However, the road's initial rotation varies, resulting in high initial angle values (e.g., 200\textdegree or 300\textdegree). Since the absolute angle values are irrelevant as long as the road remains unchanged, only the angular changes between segments are meaningful. Figure \ref{fig:raw_features_of_the_road} illustrates the angles and lengths of the first three road segments relative to the x-axis.

To avoid very large numerical values in the model's input vector and to speed up training, the angle value of the first segment is initialized to zero. For subsequent segments, the angle difference is calculated by subtracting the angle value of the preceding segment. This difference is then used as the angle information for the current segment. By doing so, the angle values for all segments are adjusted relative to their preceding segments. The recalculated angle values for the road coordinates, as shown in the example from Figure \ref{fig:raw_features_of_the_road}, are presented in Figure \ref{fig:adjusted_angles_of_the_road}.

\subsubsection{Implementation of the ITS4SDC model}

The LSTM model is implemented as a bidirectional LSTM layer. Unlike unidirectional LSTM, bidirectional LSTM processes the sequence from start to end, and then it reprocesses again from end to start. In other words, it not only considers past information but also incorporates future information. Applying reverse sequence processing in the context of roads, the future behavior of the road is as important as its past behavior. Because the AI Driver adjusts its speed and steering angle based on the upcoming curve conditions.

Moreover, it has been observed that bidirectional LSTM outperforms unidirectional LSTM when working with smaller dataset sizes (e.g., 1000, 2000, 3000). However, as the dataset size increases (e.g., 8000, 9000, 10,000), unidirectional LSTM achieves the same performance as bidirectional LSTM. The conclusion drawn from the experiments is that bidirectional LSTM either provides better results or matches the performance of unidirectional LSTM, which is the primary reason for selecting bidirectional LSTM in this study.

The LSTM model architecture is optimized through experiments. The dataset contained 197 $(x,y)$ coordinates for each road, and the best performance is achieved using a single-layer LSTM model with 220 LSTM cells. Adding more layers or increasing the number of cells did not improve performance and even hindered learning. The activation function used in LSTM cell states and outputs is \enquote{tanh}, and a dense layer with a sigmoid activation function is used at the output. The other model training configurations are as follows:
\begin{itemize}
    \item Loss function: binary cross-entropy
    \item Batch size: 1024
    \item Epoch: 400
    \item Optimizer: Adam
    \item Learning rate: 0.001.
\end{itemize}

This configuration allowed predictions of \enquote{PASS} (1) and \enquote{FAIL} (0) labels as probabilities between 0 and 1.

\begin{figure}[htbp]
    \centering
    \includegraphics[width=0.35\textwidth]{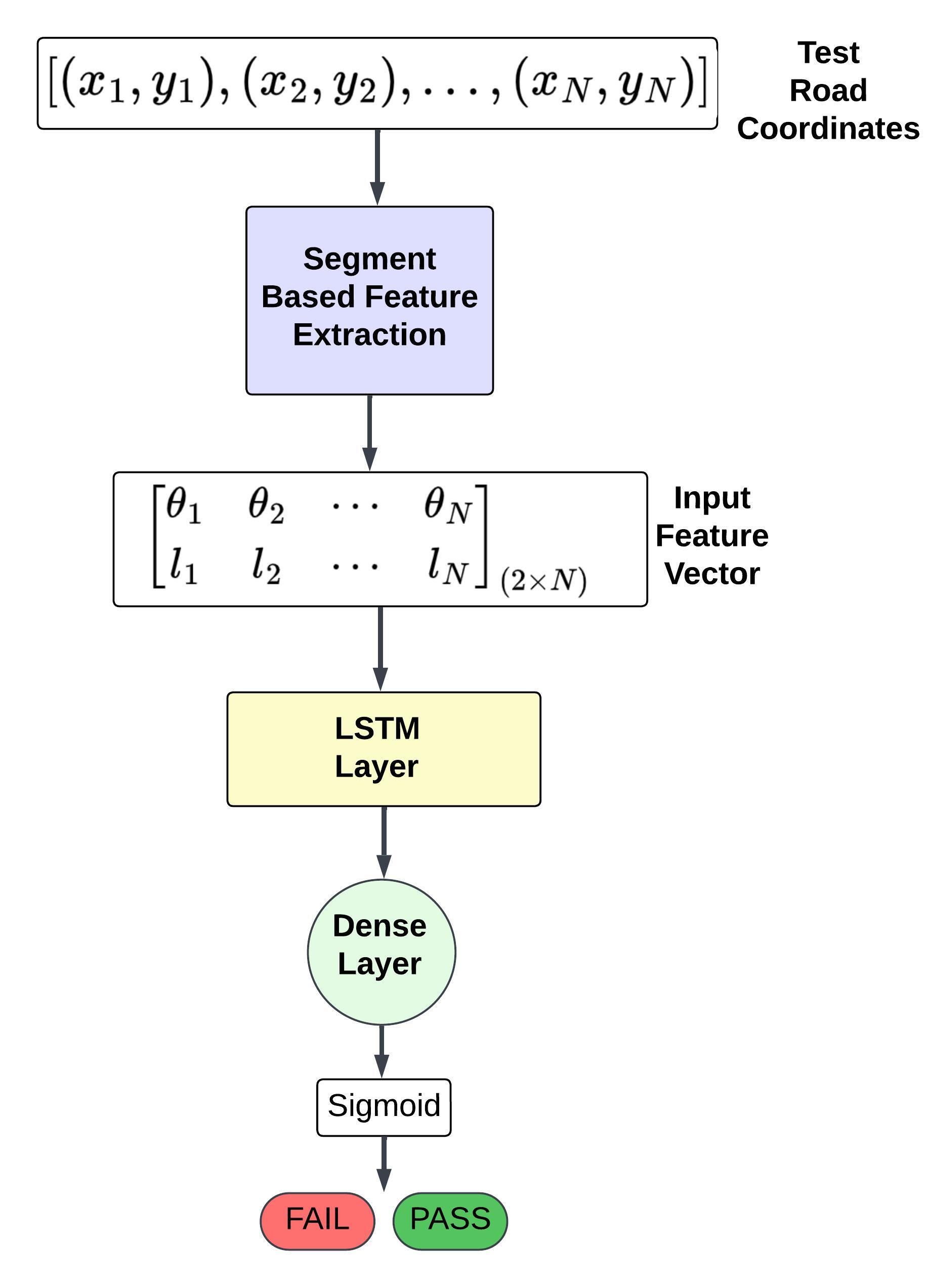}
    \caption{The flowchart of the proposed model}
    \label{fig:its4sdc_flowchart}
\end{figure}

The flowchart illustrating the working mechanism of the model is presented in Figure \ref{fig:its4sdc_flowchart}. The model takes test road coordinates as input and, in the \enquote{Segment-Based Feature Extraction} section, computes the input features required for the LSTM layer. It calculates the angle values as the angular difference between each segment and the preceding segment. At the output of the LSTM layer, a dense layer reduces the information flow from the LSTM layer to a single value. The final sigmoid layer outputs the result as a \enquote{probability} between 0 and 1; if the output value of the dense layer is greater than 0.5, the test case is interpreted as \enquote{PASS}; otherwise, it is interpreted as \enquote{FAIL}.

\section{Evaluation of the ITS4SDC Model}

The performance of the proposed model is evaluated by comparing it with SDC-Scissor \cite{b4}, a machine learning-based model developed specifically for lane-keeping systems for SDCs. 

Table \ref{table_dataset_setups} shows the setups that are used to compare ITS4SDC with SDC-Scissor. Each of the models was developed using different datasets, Dataset-1 and Dataset-2, respectively. In order to make the comparison between the two models fair, both models are trained and validated on each dataset during the evaluation (Setup 1 and Setup 2). Evaluating the model across different datasets ensures that potential bias effects or overfitting to a single dataset are minimized. For both models and in both setups 10-fold cross-validation is performed. 

Dataset-1, containing 10,000 data samples with 3,853 \enquote{FAIL} labels, has already been explained in Section \ref{subsection_dataset_preparation}. Dataset-2 was generated using the same road generation algorithm as Dataset-1 and executed in BeamNG.tech with identical simulation parameters (RF = 1.5, OOB = 0.5, MaxSpeed = 120). Dataset-2 includes 5,971 test cases, of which 3,559 test cases are valid. Among valid test cases, 1,334 test cases are labeled as \enquote{FAIL}. 

In the evaluation the following metrics are used: accuracy, precision, recall, and F1-score.

\begin{table}[htbp]
\caption{Setups for Evaluating the Models}
\begin{center}
\resizebox{0.45\textwidth}{!}{%
\begin{tabular}{|c|c|c|c|c|}
\hline
& \multicolumn{2}{|c|}{\textbf{ITS4SDC}} & \multicolumn{2}{|c|}{\textbf{SDC-Scissor}} \\
\cline{2-5}
\hline
Setup & \multicolumn{2}{|c|}{\textbf{Trained and Validated On}} & \multicolumn{2}{|c|}{\textbf{Trained and Validated On}} \\
\cline{2-5}
\hline
1 & \multicolumn{2}{|c|}{Dataset-1} & \multicolumn{2}{|c|}{Dataset-1} \\
\hline
2 & \multicolumn{2}{|c|}{Dataset-2} & \multicolumn{2}{|c|}{Dataset-2} \\
\hline
\end{tabular}
}
\label{table_dataset_setups}
\end{center}
\end{table}

\section{Results}\label{section_results}

Table \ref{table_performance_comparison_results} shows the results of the comparison between ITS4SDC and SDC-Scissor. Comparisons A to D refer to four possible combinations of setups as defined in Table \ref{table_dataset_setups}. The performance metrics show the differences that result when subtracting the average metric value of SDC-Scissor from the corresponding metric value of ITS4SDC. As mentioned in Section \ref{section_related_work}, SDC-Scissor utilizes six different machine learning models and calculates the accuracy, precision, recall, and F1 score values for each model. The primary reason for taking the average metric values of the six machine learning models in SDC-Scissor is that none of the models consistently outperformed the others. 

The ITS4SDC model demonstrated superior accuracy and precision across all comparisons (A-D). The F1 score of ITS4SDC is better in comparisons A and B, and at the same level as SDC-Scissor in comparisons C and D. The recall of ITS4SDC is better or equal in comparison A to C but lower than that of SDC-Scissor in comparison D. These results show that ITS4SDC is clearly better than SDC-Scissor when trained and validated on Dataset-1. This result might imply that ITS4SDC is overfitting on Dataset-1. Regarding SDC-Scissor results of comparisons A and B seem to indicate that it does not matter on which dataset it is trained and validated. It is even surprising that recall seems to be lower when trained and validated on Dataset-2. 

When looking at the comparisons C and D, one can see that the performance of ITS4SDC seems to be compatible with that of SDC-Scissor. Keeping in mind that ITS4SDC is a neural network model and the fact that Dataset-1 is much larger than Dataset-2 one could argue that instead of overfitting the reason for the good performance of ITS4SDC in comparisons A and B is due to the large size of Dataset-1 used for training. Neural networks tend to perform better when large training datasets are used. The difference in size between Dataset-1 and Dataset-2 might also explain why the recall of SDC-Scissor drops when using Dataset-2.

The absolute values of the performance metrics for both models are presented in Tables \ref{table_sdc_scissor_setup_1_results}, \ref{table_sdc_scissor_setup_2_results}, and \ref{table_its4sdc_results}. Tables \ref{table_sdc_scissor_setup_1_results} and \ref{table_sdc_scissor_setup_2_results} show the performance metrics of SDC-Scissor when using Dataset-1 and Dataset-2, respectively. For each performance metric, the best values are shown in italics. One can see that no model is the best on all performance metrics no matter what dataset is used.

In Table \ref{table_its4sdc_results}, Setups 1 and 2 correspond to the definitions given in Table \ref{table_dataset_setups}. Setup 3 corresponds to the situation where the ITS4SDC model is trained with the full Dataset-1 and validated on Dataset-2. Setup 4 corresponds to the situation where the ITS4SDC model is trained with the full Dataset-2 and validated on Dataset-1. Note that in Setups 1 and 2, since we do k-fold cross validation, never the full dataset is used during training. 

The results in Table \ref{table_its4sdc_results} show that Setup 1 has the best performance. This is to be expected. The drop in performance in the other setups can be explained by the use of Dataset-2 either as a training dataset or a validation dataset. 

Overall it seems that ITS4SDC is competitive with the current state of the art model SDC-Scissor with the potential to achieve clearly better performance when trained on larger dataset. The good precision achieved by ITS4SDC makes simulation-based testing of SDCs more efficient because it minimizes the number of false positives (\enquote{FAIL} mislabeled as \enquote{PASS}) and avoids executing tests that do not trigger failures. 

\begin{table}[htbp]
\centering
\caption{Performance Results of the Models}
\resizebox{0.45\textwidth}{!}{%

\begin{tabular}{|c|c|c|c|c|c|c|}
\hline
& \multicolumn{2}{|c|}{\textbf{Setup}} & \multicolumn{4}{|c|}{\textbf{Performance Metrics}} \\
\cline{2-7}
\textbf{Comparison} & \textbf{ITS4SDC} & \textbf{SDC-Scissor} & \textbf{Accuracy} & \textbf{Precision} & \textbf{Recall} & \textbf{F1} \\
\hline
A & 1 & 1 & 0.26 & 0.23 & 0.13 & 0.17 \\ 
\hline
B & 1 & 2 & 0.26 & 0.24 & 0.09 & 0.17 \\ 
\hline
C & 2 & 1 & 0.02 & 0.03 & 0.00 & 0.00 \\ 
\hline
D & 2 & 2 & 0.02 & 0.03 & -0.05 & 0.00 \\ 
\hline
\end{tabular}%
}

\label{table_performance_comparison_results}
\end{table}

\begin{table}[htbp]
\centering
\caption{SDC Scissor Setup 1 Performance Metrics}
\resizebox{0.45\textwidth}{!}{%
\begin{tabular}{|c|c|c|c|c|}
\hline
\textbf{Model} & \textbf{Accuracy} & \textbf{Precision} & \textbf{Recall} & \textbf{F1} \\
\hline
Random Forest & 0.62 & 0.65 & 0.8 & 0.72 \\ 
\hline
Gradient Boosting & \textit{0.63} &  0.64 & \textit{0.88 }& \textit{0.74 }\\ 
\hline
Support Vector Machine & \textit{0.63} & 0.65 & 0.86 &\textit{ 0.74} \\ 
\hline
Gaussian Naive Bayes & 0.57 & \textit{0.66 }& 0.6 & 0.63 \\ 
\hline
Logistic Regression & \textit{0.63} & 0.65 & 0.85 & \textit{0.74} \\ 
\hline
Decision Tree & 0.57 & 0.65 & 0.63 & \textit{0.74} \\ 
\hline
\textbf{Average Values} & \textbf{0.61} & \textbf{0.65} & \textbf{0.77} & \textbf{0.72} \\
\hline
\end{tabular}
}
\label{table_sdc_scissor_setup_1_results}
\end{table}

\begin{table}[htbp]
\centering
\caption{SDC Scissor Setup 2 Performance Metrics}
\resizebox{0.45\textwidth}{!}{%
\begin{tabular}{|c|c|c|c|c|}
\hline
\textbf{Model} & \textbf{Accuracy} & \textbf{Precision} & \textbf{Recall} & \textbf{F1} \\
\hline
Random Forest & 0.61 & 0.65 & 0.81 & 0.72 \\ 
\hline
Gradient Boosting & 0.62 &  0.64 & 0.89 & 0.74 \\ 
\hline
Support Vector Machine & \textit{0.63 }& 0.64 & \textit{0.91} & \textit{0.75} \\ 
\hline
Gaussian Naive Bayes & 0.6 & \textit{0.66 }& 0.75 & 0.7 \\ 
\hline
Logistic Regression & \textit{0.63} & 0.64 & 0.9 &\textit{ 0.75} \\ 
\hline
Decision Tree & 0.55 & 0.64 & 0.63 & 0.64 \\ 
\hline
\textbf{Average Values} & \textbf{0.61} & \textbf{0.65} & \textbf{0.82} & \textbf{0.72} \\
\hline
\end{tabular}
}
\label{table_sdc_scissor_setup_2_results}
\end{table}

\begin{table}[htbp]
    \centering
    \caption{ITS4SDC's Performance Metrics Across Setups}
    \resizebox{0.35\textwidth}{!}{
    \begin{tabular}{|c|c|c|c|c|}
         \hline   \textbf{Setup} & \textbf{Accuracy} & \textbf{Precision} & \textbf{Recall} & \textbf{F1} \\
         \hline    1 & 0.87 & 0.88 & 0.9 & 0.89  \\
         \hline    2 & 0.63 & 0.68 & 0.77 & 0.72 \\
         \hline    3 & 0.63 & 0.71 & 0.67 & 0.69 \\
         \hline    4 & 0.64 & 0.66 & 0.85 & 0.74 \\
        \hline
    \end{tabular}
    }
    \label{tab:my_label}
\label{table_its4sdc_results}
\end{table}

\section{Threats to Validity}
\label{section_threats}

This research focuses on the road classification of SDCs for lane-keeping systems. Although it demonstrates comparable performance with the current state-of-art model, several threats to validity must be acknowledged.

\textbf{Internal Validity:} The model may suffer from overfitting as only two different datasets are used for training and validation without sufficient external evaluation. The drop in recall observed during the performance comparison between ITS4SDC and SDC-Scissor highlights one of the limitations of the ITS4SDC model. Additionally,  one major drawback of applying deep learning-based models such as LSTM is the requirement for fixed input feature dimensions. The experiments indicated that the ITS4SDC model is sensitive to parameters such as the activation function and the number of layers. Regarding the road data, input sizes can be adjusted for the LSTM model using interpolation to increase dimensions or sampling to reduce them. The selection of too many points along the road through interpolation creates an obstacle for the model's learning as it leads to a decrease in segment lengths and segment angles. However, this issue can be avoided by sampling the points along the road and selecting one out of every two or three points. For the dataset used in the competition, the 197 interpolated road coordinates contain learnable segment angle and length data for the model, so no further interpolation or alternative sampling algorithms were deemed necessary. 

\textbf{External Validity:} The experiments were conducted in the BeamNG simulation environment using an ETK-800 vehicle on a flat 2D asphalt surface with no slope. Although the BeamNG simulator features soft-body dynamics and can closely model real-world conditions, there may still be discrepancies between the real world and the simulator that should be taken into account. One of the main reasons a vehicle leaves the lane is its dynamic and electronic properties. Therefore, failure-inducing roads vary from vehicle to vehicle, and the dataset used to train the model should be collected based on the target vehicle.   

\textbf{Construct Validity:} The binary classification of \enquote{safe} and \enquote{unsafe} simplifies the problem, but may overlook the borderline cases that could be critical for safety evaluations. The OOB parameter defines the extent (as a percentage) of the vehicle’s deviation from the road or lane markings and is used to define the tolerance that classifies roads that are used as test cases into \enquote{safe} and \enquote{unsafe}. For example, a tolerance of 100\% means that a road is only classified as \enquote{unsafe} if the vehicle has completely left the road. While a tolerance of 0\% means that a road is classified as \enquote{unsafe} if less than 1\% of the vehicle left the road.


For the development of the ITS4SDC model the OOB tolerance was set to 50\% in conformance with the requirements of the competition. Experiments showed that changing the OOB tolerance in the range 20\% to 100\% did not change the labels of the test cases. One reason for this seems to be the slippage of the vehicle on the surface outside the road for any OOB tolerance bigger than 20\%. Conversely, experiments indicated that when only less than 20\% of the vehicle's size left the road, the vehicle often managed to re-enter the road, classifying the road as \enquote{safe}. Therefore, we assume that the accuracy of our model, which was trained using the 50\% OOB tolerance, might decrease if the OOB tolerance is reduced to a value smaller than 20\%. 

The vehicle's electronic controllers also play an important role. The electronic controllers of the ETK-800, such as the Electronic Stability Controller (ESC) and Anti-lock Braking System (ABS), which activate when the vehicle is about to go off the road and do not always help maintain control in certain situations, might make it easier to keep the vehicle on the road in a different vehicle. Therefore, it is clear that the dataset should be collected specifically for the vehicle. 

Another factor contributing to the vehicle's off-road departure is the risk factor, which has been set to \enquote{moderate driver} (1.5) as used in the competition. Increasing the risk factor leads the vehicle to ignore turns more often, resulting in more fail roads in the dataset. On the other hand, setting the risk factor too low causes the vehicle to avoid taking risks, maintaining a very low speed on all roads, and completing them. No study has been conducted on the model's sensitivity to changes in the risk factor, and its impact on classification performance is unknown.



\section{Conclusions}\label{section_conclusion}

An LSTM-based approach to the test selection problem for SDCs, specifically focusing on lane-keeping systems, has been presented in this paper. By treating road data as sequences and utilizing an LSTM-based model, roads were classified as \enquote{SAFE} or \enquote{UNSAFE}. The proposed model was comprehensively evaluated using two different datasets to assess its generalization capability. It demonstrated superior performance in accuracy and precision metrics compared to machine learning models while maintaining competitive performance in recall and F1 scores across both datasets. These findings highlight the potential of deep learning methods, particularly sequence-based models, in addressing road classification problems for SDCs.

In the presented work, road coordinates were generated using the Frenetic algorithm. Evaluating the model on datasets generated by other road generation algorithms such as Deeper\cite{b19} and Swat\cite{b20} could be interesting future work. Considering that the most critical road feature causing vehicles to leave the road is related to the curve regions, future work might explore attention-based LSTM algorithms by treating these curve regions as \enquote{attention} zones to improve model performance.

\section{Acknowledgements}

This study was co-funded by the European Union and the Estonian Ministry of Education and Research via project TEM-TA120, by BMK, BMAW, and the State of Upper Austria in the frame of the SCCH competence center INTEGRATE [(FFG grant no. 892418)] part of the FFG COMET Competence Centers for Excellent Technologies Programme, by grant PRG1226 of the Estonian Research Council, and by Bolt Technology ÖU.

We gratefully acknowledge the BeamNG company for providing us with the simulation environment that enabled the test case execution and dataset preparation.

\end{document}